\pdfoutput=1

\documentclass[final]{cvpr}

\usepackage{times}
\usepackage{epsfig}
\usepackage{graphicx}
\usepackage{amsmath}
\usepackage{amssymb}

\usepackage[ruled,vlined]{algorithm2e}

\usepackage{times}
\usepackage{epsfig}
\usepackage{graphicx}
\usepackage{color}
\usepackage{epsfig}
\usepackage{comment}
\usepackage{enumitem}
\usepackage[export]{adjustbox}
\usepackage{tabularx}
\usepackage[table]{xcolor}
\usepackage{colortbl}
\usepackage{subfig}
\usepackage{multirow}
\usepackage{booktabs}
\usepackage{xr}
\usepackage{pifont}
\usepackage[hyphens]{url}
\usepackage{sidecap}
\usepackage{tablefootnote}

\makeatletter
  \newcommand\figcaption{\def\@captype{figure}\caption}
  \newcommand\tabcaption{\def\@captype{table}\caption}
\makeatother

\newcommand{\Paragraph}[1]{\vspace{1mm} \noindent \textbf{#1} \hspace{0mm}}
\newcommand{\Section}[1]{\vspace{-1mm} \section{#1} \vspace{-1mm}}

\definecolor{Gray}{gray}{0.93}
\newcommand{\kevin}[1]{\textcolor{black}{#1}}
\newcommand{\lijuanw}[1]{\textcolor{black}{#1}}

\newcommand{\kevinarxiv}[1]{\textcolor{black}{#1}}
\newcommand{\camready}[1]{\textcolor{black}{#1}}

\usepackage[pagebackref=true,breaklinks=true,colorlinks,bookmarks=false]{hyperref}

\def\cvprPaperID{1740} 
\def\confYear{CVPR 2021}

\begin{document}

\title{End-to-End Human Pose and Mesh Reconstruction with Transformers}

\author{Kevin Lin \ \ \  Lijuan Wang \ \ \ Zicheng Liu\\
Microsoft\\
{\tt\small \{keli, lijuanw, zliu\}@microsoft.com}
}

\maketitle

\begin{abstract}
    
   We present a new method, called MEsh TRansfOrmer (METRO), to reconstruct 3D human pose and mesh vertices from a single image. Our method uses a transformer encoder to jointly model \lijuanw{vertex-vertex and vertex-joint} 
   interactions, and outputs 3D joint coordinates and mesh vertices simultaneously. Compared to existing techniques that regress pose and shape parameters, METRO does not rely on any parametric mesh models like SMPL, thus it can be easily extended to other objects such as hands. We further relax the mesh topology and allow the transformer self-attention mechanism to freely attend between any two vertices, making it possible to learn non-local relationships among mesh vertices and joints. With the \lijuanw{proposed} masked vertex modeling, our method is more robust and effective in handling challenging situations like partial occlusions. METRO generates new state-of-the-art results for human mesh reconstruction on the public Human3.6M and 3DPW datasets. Moreover, we demonstrate the generalizability of METRO to 3D hand reconstruction in the wild, \kevinarxiv{outperforming existing state-of-the-art methods on FreiHAND dataset.} Code and pre-trained models are available at \url{https://github.com/microsoft/MeshTransformer}.

\end{abstract}

\pdfoutput=1
\section{Introduction}

3D human pose and mesh reconstruction from a single image has attracted a lot of attention because it has many applications including virtual reality, sports motion analysis, neurodegenerative condition diagnosis, etc. It is a challenging problem due to complex articulated motion and occlusions. 

Recent work in this area can be roughly divided into two categories. Methods in the first category use a parametric model like SMPL~\cite{loper2015smpl} and learn to predict shape and \kevin{pose} coefficients~\cite{guan2009estimating,lassner2017unite,pavlakos2018learning,kanazawa2018end,kolotouros2019learning,omran2018neural,Rong_2019_ICCV,kocabas2019vibe}. 
\kevin{Great success has been achieved with this approach. The strong prior encoded in the parametric model increases its robustness to environment variations.}
A drawback of this approach is that the pose and shape spaces are constrained by the limited exemplars that are used to construct the parametric model. To overcome this limitation, methods in the second category do not use any parametric models~\cite{kolotouros2019convolutional,Choi_2020_ECCV_Pose2Mesh,Moon_2020_ECCV_I2L-MeshNet}. These methods either use a graph convolutional neural network to model neighborhood vertex-vertex interactions~\cite{kolotouros2019convolutional,Choi_2020_ECCV_Pose2Mesh}, or use 1D heatmap to regress vertex coordinates~\cite{Moon_2020_ECCV_I2L-MeshNet}. One limitation with these approaches is that they are not efficient in modeling non-local vertex-vertex interactions. 

\begin{figure}[t]
\begin{center}
\includegraphics[trim=0 0 448 0, clip,width=0.32\columnwidth]{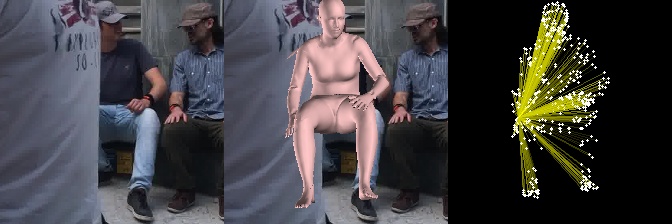}
\includegraphics[trim=448 0 0 0, clip,width=0.32\columnwidth]{fig/results_3dpw_batch150_jpg_att.jpg}
\includegraphics[width=0.32\columnwidth]{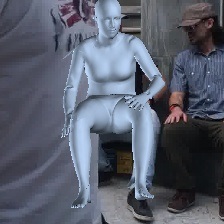}
\setlength{\tabcolsep}{33.0pt}
\begin{tabular}{ccc}
(a) & (b) & (c)\\
\end{tabular}
\caption{
METRO learns non-local interactions among body joints and mesh vertices for human mesh reconstruction. Given an input image in (a), METRO predicts human mesh by taking non-local interactions into consideration. (b) illustrates the attentions between the occluded wrist joint and the mesh vertices where brighter color indicates stronger attention. (c) is the reconstructed mesh.
} 
\vspace{-0mm}
\label{fig:fig1}
\end{center}
\end{figure}

Researchers have shown that there are strong correlations between non-local vertices which may belong to different parts of the body (e.g. hand and foot)~\cite{wang2012mining}. 
\kevin{In computer graphics and robotics, inverse kinematics techniques~\cite{aristidou2018inverse} have been developed to estimate the internal joint positions of an articulated figure given the position of an end effector such as a hand tip. We believe that learning the correlations among body joints and mesh vertices including both short range and long range ones is valuable for handling challenging poses and occlusions in body shape reconstruction.}
In this paper, we propose a simple yet effective framework to model global vertex-vertex interactions. The main ingredient of our framework is a transformer.

Recent studies show that transformer~\cite{vaswani2017attention} significantly improves the performance on various tasks in natural language processing~\cite{bahdanau2015neural,devlin2019bert,radford2018improving,radford2019language}. The success is mainly attributed to the \lijuanw{self-}attention mechanism of a transformer, which is particularly effective in modeling the dependencies (or interactions) without regard to their distance in both inputs and outputs. Given the dependencies, transformer is able to \textit{soft-search} the relevant tokens and performs prediction based on the important features~\cite{bahdanau2015neural,vaswani2017attention}.

In this work, \lijuanw{we propose METRO, a multi-layer Transformer encoder with progressive dimensionality reduction,}
to reconstruct 3D body joints and mesh vertices from a given input image, simultaneously. 
We design the Masked Vertex Modeling objective with a transformer encoder architecture to \lijuanw{enhance}
the interactions among joints and vertices. As shown in Figure~\ref{fig:fig1}, METRO learns to discover both short- and long-range interactions among body joints and mesh vertices, which helps to better reconstruct the 3D human body shape with large pose variations and occlusions. 

Experimental results on multiple public datasets demonstrate that 
METRO is effective in learning vertex-vertex and vertex-joint interactions, and consequently outperforms the prior works on human mesh reconstruction by a large margin.
To the best of our knowledge, METRO is the first approach that leverages a transformer encoder architecture to jointly learn 3D human pose and mesh reconstruction from a single input image. Moreover, METRO is a general framework which can be easily applied to predict a different 3D mesh, for example, to reconstruct a 3D hand from an input image.

In summary, we make the following contributions.
\begin{itemize}
\item{We introduce a new transformer-based method, named METRO, for 3D human pose and mesh reconstruction from a single image.}
\item{We design the Masked Vertex Modeling objective with a multi-layer transformer encoder to model both vertex-vertex and vertex-joint interactions for better reconstruction.}
\item{METRO achieves new state-of-the-art performance on the large-scale benchmark Human3.6M and the challenging 3DPW dataset.}
\item{METRO is a versatile framework that can be easily realized to predict a different type of 3D mesh, such as 3D hand as demonstrated in the experiments. METRO achieves the first place on FreiHAND leaderboard \camready{at the time of paper submission}.}
\end{itemize}

\pdfoutput=1
\section{Related Works}\label{sec:related}

\Paragraph{Human Mesh Reconstruction (HMR):} HMR is a task of reconstructing 3D human body shape, which is an active research topic in recent years. While pioneer works have demonstrated impressive reconstruction using various sensors, such as depth sensors~\cite{newcombe2011kinectfusion,shin20193d} or inertial measurement units ~\cite{DIP:SIGGRAPHAsia:2018,vonMarcard2018}, researchers are exploring to use a monocular camera setting that is more efficient and convenient. However, HMR from a single image is difficult due to complex pose variations, occlusions, and limited 3D training data.

Prior studies propose to adopt the pre-trained parametric human models, \ie, SMPL~\cite{loper2015smpl}, STAR~\cite{STAR:2020}, MANO~\cite{MANO:SIGGRAPHASIA:2017}, and estimate the pose and shape coefficients of the parametric model for HMR. Since it is challenging to regress the pose and shape coefficients directly from an input image, recent works further propose to leverage various human body priors such as human skeletons~\cite{lassner2017unite,pavlakos2018learning} or segmentation maps~\cite{omran2018neural}, and explore different optimization strategies~\cite{kolotouros2019learning,kanazawa2018end,tung2017self,guan2009estimating} and temporal information~\cite{kocabas2019vibe} to improve reconstruction.

On the other hand, instead of adopting a parametric human model, researchers have also proposed approaches to directly regress 3D human body shape from an input image. For example, researchers have explored to represent human body using a 3D mesh~\cite{kolotouros2019convolutional,Choi_2020_ECCV_Pose2Mesh}, a volumetric space~\cite{varol2018bodynet}, or an occupancy field~\cite{saito2019pifu,saito2020pifuhd}. Each of the prior works addresses a specific output representation for their target application.
Among the literature, the relevant study is GraphCMR~\cite{kolotouros2019convolutional}, which aims to regress 3D mesh vertices using graph convolutional neural networks (GCNNs). Moreover, recent proposed Pose2Mesh~\cite{Choi_2020_ECCV_Pose2Mesh} is a cascaded model using GCNNs. Pose2Mesh reconstructs human mesh based on the given human pose representations.

\kevin{While GCNN-based methods~\cite{Choi_2020_ECCV_Pose2Mesh,kolotouros2019convolutional} are} designed to model neighborhood vertex-vertex interactions based on a pre-specified mesh topology, it is less efficient in modeling longer range interactions. In contrast, METRO models global interactions among joints and mesh vertices without being limited by any mesh topology. \kevin{In addition, our method learns with self-attention mechanism, which is different from prior studies~\cite{Choi_2020_ECCV_Pose2Mesh,kolotouros2019convolutional}.}

\begin{figure*}[h]
\begin{center}
\includegraphics[trim=0 0 0 0, clip,width=1\textwidth]{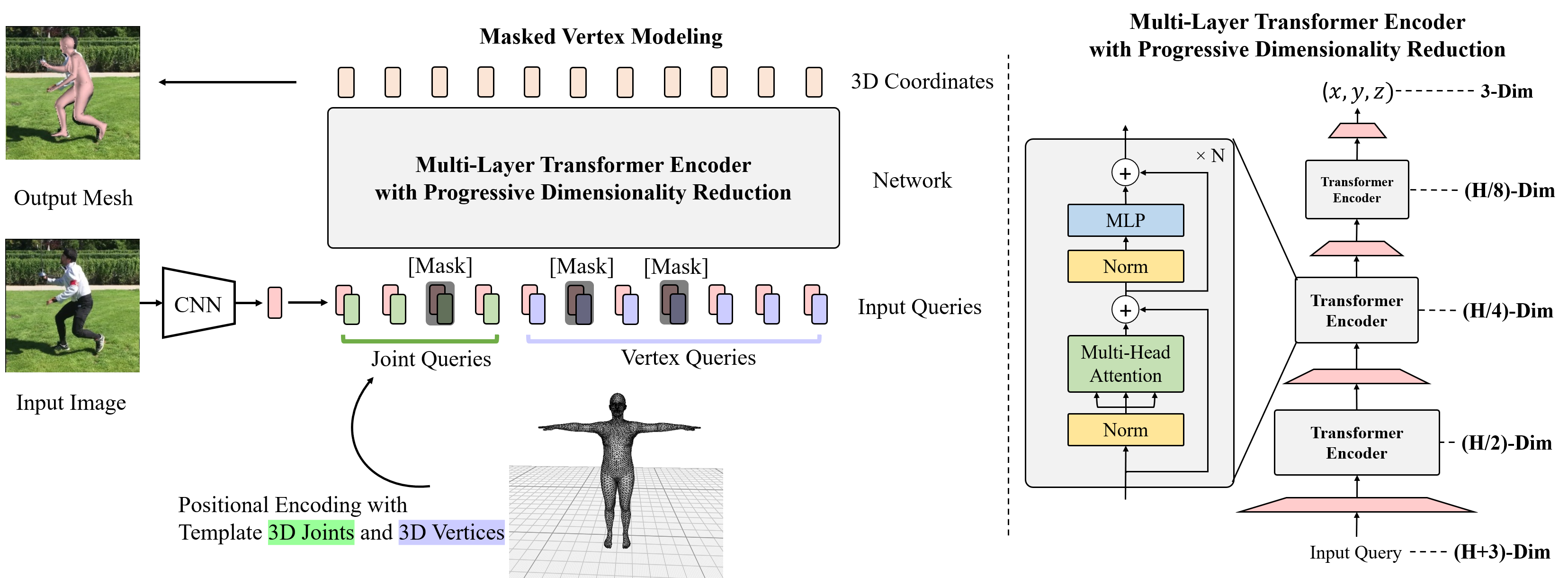}
\caption{
\textbf{Overview of the proposed framework.} Given an input image, we extract an image feature vector using a convolutional neural network (CNN). We perform position encoding by adding a template human mesh to the image feature vector by concatenating the image feature with the 3D coordinates ($x_i, y_i, z_i$) of every body joint $i$, and 3D coordinates ($x_j, y_j, z_j$) of every vertex $j$. Given a set of joint queries and vertex queries, we perform self-attentions through multiple layers of a transformer encoder, and regress the 3D coordinates of body joints and mesh vertices in parallel. We use a progressive dimensionality reduction architecture (right) to gradually reduce the hidden embedding dimensions from layer to layer. Each token in the final layer outputs 3D coordinates of a joint or mesh vertex. \kevin{Each encoder block has $4$ layers and $4$ attention heads. $H$ denotes the dimension of an image feature vector.}
} 
\vspace{-0mm}
\label{fig:overview}
\end{center}
\end{figure*}

\Paragraph{Attentions and Transformers:}
Recent studies~\cite{parikh2016decomposable,lin2016neural,vaswani2017attention} have shown that attention mechanisms improve the performance on various language tasks. Their key insight is to learn the attentions to \textit{soft-search} relevant inputs that are important for predicting an output~\cite{bahdanau2015neural}. 
Vaswani~\etal~\cite{vaswani2017attention} further propose a transformer architecture based solely on attention mechanisms.  
Transformer is highly parallelized using multi-head self-attention for efficient training and inference, and leads to superior performance in language modeling at scale, as explored in BERT~\cite{devlin2019bert} and GPT~\cite{radford2018improving,radford2019language,brown2020language}.

Inspired by the recent success in neural language field, there is a growing interest in exploring the use of transformer architecture for various vision tasks, such as learning the pixel distributions for image generation~\cite{chen2020generative,parmar2018image} and classification~\cite{chen2020generative,dosovitskiy2020image}, or to simplify object detection as a set prediction problem~\cite{carion2020end}. However, 3D human reconstruction has not been explored along this direction.

\kevin{In this study, we present a multi-layer transformer architecture with progressive dimensionality reduction to regress the 3D coordinates of the joints and vertices.}

\pdfoutput=1

\section{Method}\label{sec:method}

Figure~\ref{fig:overview} is an overview of our proposed framework. It takes an image of size $224 \times224$ as input, and predicts a set of body joints $J$ and mesh vertices $V$. The proposed framework consists of two modules: \textit{Convolutional Neural Network}, and \textit{Multi-Layer Transformer Encoder}. First, we use a CNN to extract an image feature vector from an input image. Next, Multi-Layer Transformer Encoder takes as input the feature vector and outputs the 3D coordinates of the body joint and mesh vertex in parallel. We describe each module in details as below.

\subsection{Convolutional Neural Network}
In the first module of our framework, we employ a Convolutional Neural Network (CNN) for feature extraction. The CNN is pre-trained on ImageNet classification task~\cite{russakovsky2015imagenet}. Specifically, we extract a feature vector $X$ from the last hidden layer. The extracted feature vector $X$ is typically of dimension $2048$. We input the feature vector $X$ to the transformer for the regression task.  

With this generic design, it allows an end-to-end training for human pose and mesh reconstruction. Moreover, transformer can easily benefit from large-scale pre-trained CNNs\kevin{, such as HRNets~\cite{WangSCJDZLMTWLX19}.} In our experiments, we conduct analysis on the input features, and discover that high-resolution image features are beneficial for transformer to regress 3D coordinates of body joints and mesh vertices.

\subsection{Multi-Layer Transformer Encoder with Progressive Dimensionality Reduction}

Since we need to output 3D coordinates, we cannot directly apply the existing transformer encoder architecture~\cite{dosovitskiy2020image,carion2020end} because they use a constant dimensionality of the hidden embeddings for all the transformer layers. \camready{Inspired by~\cite{hinton2006reducing} which performs dimentionality reduction gradually with multiple blocks, we design a new architecture with a progressive dimensionality reduction scheme.} As shown in Figure~\ref{fig:overview} right, we use linear projections to reduce the dimensionality of the hidden embedding after each encoder layer. By adding multiple encoder layers, the model is viewed as performing self-attentions and dimensionality reduction in an alternating manner. The final output vectors of our transformer encoder are the 3D coordinates of the joints and mesh vertices.

As illustrated in Figure~\ref{fig:overview} left, the input to the transformer encoder are the body joint and mesh vertex queries. \camready{In the same spirit as positional encoding~\cite{vaswani2017attention,kolotouros2019convolutional,groueix20183d},} we use a template human mesh to preserve the positional information of each query in the input sequence. To be specific, we concatenate the image feature vector $X\in \mathbb{R}^{2048\times1}$ with the 3D coordinates $(x_i, y_i, z_i)$ of every body joint $i$. This forms a set of joint queries $Q_J=\{q_{1}^J,q_{2}^J,\dots,q_{n}^J\}$, where $q_{i}^J \in \mathbb{R}^{2051\times1}$. Similarly, we conduct the same positional encoding for every mesh vertex $j$, and form a set of vertex queries $Q_V=\{q_{1}^V,q_{2}^V,\dots,q_{m}^V\}$, where  $q_{j}^V \in \mathbb{R}^{2051\times1}$.

\subsection{Masked Vertex Modeling}

Prior works~\cite{devlin2019bert,taylor1953cloze} use the Masked Language Modeling (MLM) to learn the linguistic properties of a training corpus. However, MLM aims to recover the inputs, which cannot be directly applied to our regression task. 

To fully activate the bi-directional attentions in our transformer encoder, we design a Masked Vertex Modeling (MVM) for our regression task. We mask some percentages of the input queries at random. Different from recovering the masked inputs like MLM~\cite{devlin2019bert}, we instead ask the transformer to regress all the joints and vertices. 

In order to predict an output corresponding to a missing query, the model will have to resort to other relevant queries. This is in spirit similar to simulating occlusions where partial body parts are invisible. As a result, MVM enforces transformer to regress 3D coordinates by taking other relevant vertices and joints into consideration, without regard to their distances and mesh topology.
This facilitates both short- and long-range interactions among joints and vertices for better human body modeling.

\subsection{Training}

To train the transformer encoder, we apply loss functions on top of the transformer outputs, and minimize the errors between predictions and ground truths. Given a dataset $D=\{I^i, \bar{V}_{3D}^i, \bar{J}_{3D}^i, \bar{J}_{2D}^i\}_{i=1}^{T}$, where $T$ is the total number of training images. $I \in \mathbb{R}^{w\times h \times 3}$ denotes an RGB image. $\bar{V}_{3D} \in \mathbb{R}^{M \times 3}$ denotes the ground truth 3D coordinates of the mesh vertices and $M$ is the number of vertices. $\bar{J}_{3D} \in \mathbb{R}^{K \times 3}$ denotes the ground truth 3D coordinates of the body joints and $K$ is the number of joints of a person. Similarly, $\bar{J}_{2D} \in \mathbb{R}^{K \times 2}$ denotes the ground truth 2D coordinates of the body joints. 

Let $V_{3D}$ denote the output vertex locations, and $J_{3D}$ is the output joint locations, we use $L_1$ loss to minimize the errors between predictions and ground truths:\begin{equation}\begin{aligned}%
\label{eqn:vertex-loss}%
\mathcal{L}_{V} = \frac{1}{M}\sum_{i=1}^{M} \left| \left| V_{3D}-\bar{V}_{3D} \right| \right|_1,
\end{aligned}
\end{equation}
\begin{equation}\begin{aligned}%
\label{eqn:3dpose-loss}%
\mathcal{L}_{J} = \frac{1}{K}\sum_{i=1}^{K} \left| \left| J_{3D}-\bar{J}_{3D} \right| \right|_1.
\end{aligned}
\end{equation}

It is worth noting that, the 3D joints can also be calculated from the predicted mesh. Following the common practice in literature~\cite{Choi_2020_ECCV_Pose2Mesh,kanazawa2018end,kolotouros2019convolutional,kolotouros2019learning}, we use a pre-defined regression matrix $G \in \mathbb{R}^{K \times M}$, and obtain the regressed 3D joints by $J_{3D}^{reg} = GV_{3D}$. Similar to prior works, we use $L_1$ loss to optimize $J_{3D}^{reg}$:   
\begin{equation}\begin{aligned}%
\label{eqn:3dpose-reg-loss}%
\mathcal{L}_{J}^{reg} = \frac{1}{K}\sum_{i=1}^{K} \left| \left| J_{3D}^{reg}-\bar{J}_{3D} \right| \right|_1.
\end{aligned}
\end{equation}

2D re-projection has been commonly used to enhance the image-mesh alignment~\cite{kanazawa2018end,kolotouros2019convolutional,kolotouros2019learning}. Also, it helps visualize the reconstruction in an image. Inspired by the prior works, we project the 3D joints to 2D space using the estimated camera parameters, and minimize the errors between the 2D projections and 2D ground truths:

\begin{equation}
\begin{aligned}%
\label{eqn:2dpose-loss}%
\mathcal{L}_{J}^{proj} = \frac{1}{K}\sum_{i=1}^{K} \left| \left| J_{2D}-\bar{J}_{2D} \right| \right|_1,
\end{aligned}
\end{equation}
where the camera parameters are learned by using a linear layer on top of the outputs of the transformer encoder.

To perform large-scale training, it is highly desirable to leverage both 2D and 3D training datasets for better generalization. As explored in literature~\cite{omran2018neural,kanazawa2018end,kolotouros2019convolutional,kolotouros2019learning,kocabas2019vibe,Choi_2020_ECCV_Pose2Mesh,Moon_2020_ECCV_I2L-MeshNet}, we use a mix-training strategy that leverages different training datasets, with or without the paired image-mesh annotations. Our overall objective is written as:
\begin{equation}\begin{aligned}%
\label{eqn:all-loss}%
\mathcal{L} = \alpha\times(\mathcal{L}_{V} + \mathcal{L}_{J} + \mathcal{L}_{J}^{reg}) + \beta \times \mathcal{L}_{J}^{proj},
\end{aligned}
\end{equation}
where $\alpha$ and $\beta$ are binary flags for each training sample, indicating the availability of 3D and 2D ground truths, respectively.

\subsection{Implementation Details}

Our method is able to process arbitrary sizes of mesh. However, due to memory constraints of current hardware, \camready{our transformer processes a coarse mesh: (1) We use a coarse template mesh (431 vertices) for positional encoding, and transformer outputs a coarse mesh; (2) We use learnable Multi-Layer Perceptrons (MLPs) to upsample the coarse mesh to the original mesh (6890 vertices for SMPL human mesh topology); (3) The transformer and MLPs are trained end-to-end; Please note that the coarse mesh is obtained by sub-sampling twice to $431$ vertices with a sampling algorithm~\cite{ranjan2018generating}. As discussed in the literature~\cite{kolotouros2019convolutional}, the implementation of learning a coarse mesh followed by upsampling is helpful to reduce computation. It also helps avoid redundancy in original mesh (due to spatial locality of vertices), which makes training more efficient.}

\pdfoutput=1
\begin{table*}[h]
\centering
\begin{tabular}{lcccccc}
    \toprule
    \multirow{1}{*}{} & \multicolumn{3}{c}{3DPW} & & \multicolumn{2}{c}{Human3.6M}\\ 
    \cline{2-4}\cline{6-7}
	Method  & MPVE $\downarrow$ & MPJPE $\downarrow$ & PA-MPJPE $\downarrow$ & &  MPJPE $\downarrow$ & PA-MPJPE $\downarrow$ \\
	\midrule
	HMR~\cite{kanazawa2018end} & $-$ & $-$ & $81.3$ && $88.0$ & $56.8$  \\
	GraphCMR~\cite{kolotouros2019convolutional} & $-$ & $-$ & $70.2$ && $-$ & $50.1$\\
	SPIN~\cite{kolotouros2019learning} & $116.4$ & $-$ & $59.2$ && $-$ & $41.1$\\
	Pose2Mesh~\cite{Choi_2020_ECCV_Pose2Mesh} & $-$ & $89.2$ & $58.9$ && $64.9$ & $47.0$\\
	I2LMeshNet~\cite{Moon_2020_ECCV_I2L-MeshNet} & $-$ & $93.2$ & $57.7$ && $55.7$ & $41.1$\\
	VIBE~\cite{kocabas2019vibe} & $99.1$ & $82.0$ & $51.9$ && $65.6$ & $41.4$\\
	\midrule
    METRO (Ours)  & $\textbf{88.2}$ & $\textbf{77.1}$ & $\textbf{47.9}$ && $\kevin{\textbf{54.0}}$ & $\kevin{\textbf{36.7}}$\\ 
	\bottomrule
\end{tabular}
\caption{Performance comparison with the state-of-the-art methods on 3DPW and Human3.6M datasets.
}
\vspace{0mm}
\label{tbl:compare-h36m-3dpw}
\end{table*}

\section{Experimental Results}\label{sec:exp}

We first show that our method outperforms the previous state-of-the-art human mesh reconstruction methods on Human3.6M and 3DPW datasets. Then, we provide ablation study and insights for the non-local interactions and model design. Finally, we demonstrate the generalizability of our model on hand reconstruction.

\subsection{Datasets}

Following the literature~\cite{omran2018neural,kanazawa2018end,kolotouros2019convolutional,kolotouros2019learning,kocabas2019vibe,Choi_2020_ECCV_Pose2Mesh,Moon_2020_ECCV_I2L-MeshNet}, we conduct mix-training using 3D and 2D training data. We describe each dataset below.

\Paragraph{Human3.6M~\cite{ionescu2014human3}} is a large-scale dataset with 2D and 3D annotations. Each image has a subject performing a different action. Due to the license issue, the groundtruth 3D meshes are not available. Thus, we use the pseudo 3D meshes provided in~\cite{Choi_2020_ECCV_Pose2Mesh,Moon_2020_ECCV_I2L-MeshNet} for training. The pseudo labels are created by model fitting with SMPLify-X~\cite{SMPL-X:2019}. For evaluation, we use the groundtruth 3D pose labels provided in Human3.6M for fair comparison. Following the common setting~\cite{tekin2016direct,kolotouros2019convolutional,kanazawa2018end}, we train our models using subjects S1, S5, S6, S7 and S8. We test the models using subjects S9 and S11.

\Paragraph{\textbf{3DPW}~\cite{vonMarcard2018}} is an outdoor-image dataset with 2D and 3D annotations. The training set consists of $22K$ images, and the test set has $35K$ images. Following the previous state-of-the-arts~\cite{kocabas2019vibe}, we use 3DPW training data when conducting experiments on 3DPW. 

\Paragraph{\textbf{UP-3D~\cite{lassner2017unite}}} is an outdoor-image dataset. Their 3D annotations are created by model fitting. The training set has $7K$ images.
 
\Paragraph{\textbf{MuCo-3DHP}~\cite{mehta2018single}} is a synthesized dataset based on MPI-INF-3DHP dataset~\cite{mehta2017monocular}. It composites the training data with a variety of real-world background images. It has $200K$ training images. 

\Paragraph{\textbf{COCO~\cite{lin2014microsoft}}} is a large-scale dataset with 2D annotations. We also use the pseudo 3D mesh labels provided in~\cite{kolotouros2019learning}, which are fitted with SMPLify-X~\cite{SMPL-X:2019}. 

\Paragraph{\textbf{MPII~\cite{andriluka14cvpr}}} is an outdoor-image dataset with 2D pose labels. The training set consists of $14K$ images. 

\Paragraph{\textbf{FreiHAND~\cite{zimmermann2019freihand}}} is a 3D hand dataset. The training set consists of $130K$ images, and the test set has $4K$ images. We demonstrate the generalizability of our model on this dataset. We use the provided set for training, and conduct evaluation on their online server.

\subsection{Evaluation Metrics}

We report results using three standard metrics as below. The unit for the three metrics is millimetter (mm).

\Paragraph{\textbf{MPJPE}:} Mean-Per-Joint-Position-Error (MPJPE)~\cite{ionescu2014human3} is a metric for evaluating human 3D pose~\cite{kanazawa2018end,kolotouros2019learning,Choi_2020_ECCV_Pose2Mesh}. MPJPE measures the Euclidean distances between the ground truth joints and the predicted joints.

\Paragraph{\textbf{PA-MPJPE}:} PA-MPJPE, or Reconstruction Error~\cite{zhou2018monocap}, is another metric for this task. It first performs a 3D alignment using Procrustes analysis (PA)~\cite{gower1975generalized}, and then computes MPJPE. PA-MPJPE is commonly used for evaluating 3D reconstruction~\cite{zhou2018monocap} as it measures the errors of the reconstructed structure without regard to the scale and rigid pose (\ie, translations and rotations).

\Paragraph{\textbf{MPVE}:} Mean-Per-Vertex-Error (MPVE)~\cite{pavlakos2018learning} measures the Euclidean distances between the ground truth vertices and the predicted vertices.

\begin{figure*}[t]
\begin{center}
\includegraphics[trim=0 0 0 0, clip,width=1\textwidth]{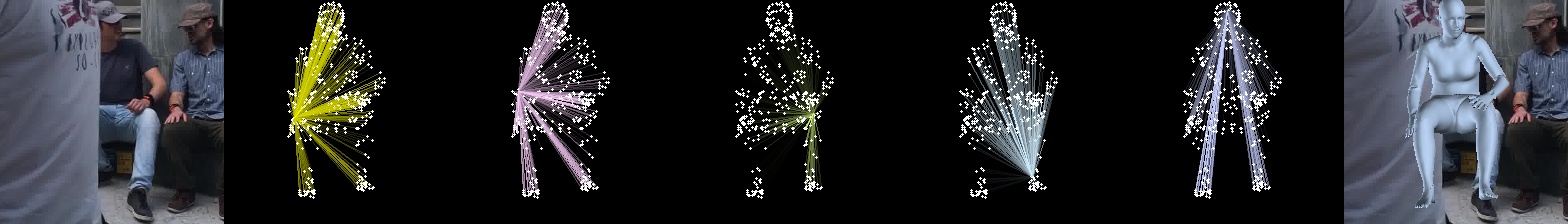}\\
\includegraphics[trim=0 0 0 0, clip,width=1\textwidth]{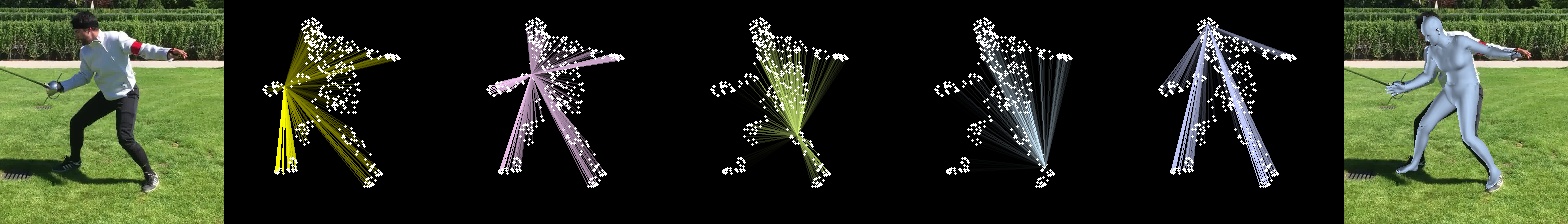}\\
\includegraphics[trim=0 0 0 0, clip,width=1\textwidth]{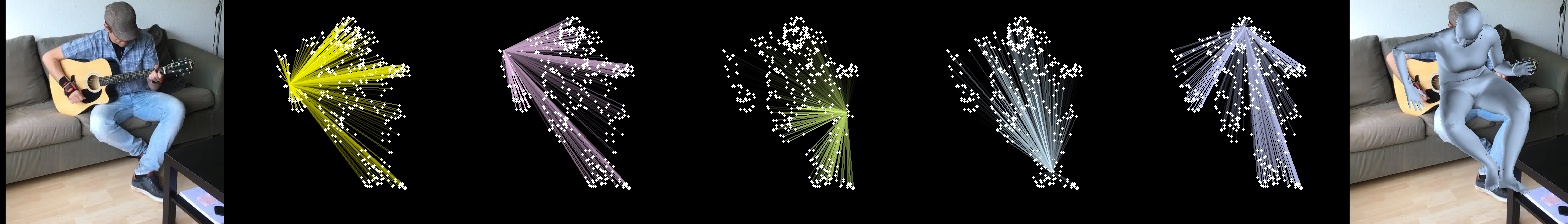}\\
\includegraphics[trim=0 0 0 0, clip,width=1\textwidth]{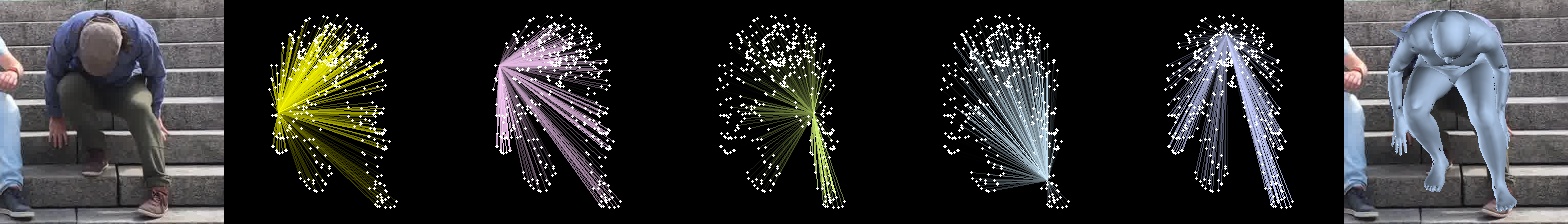}\\
\setlength{\tabcolsep}{21pt}
\begin{tabular}{ccccccc}
Input & R-Wrist &  R-Elbow &  L-Knee &  L-Ankle & Head & Output
\end{tabular}
\vspace{-2mm}
\caption{
\kevin{Qualitative results of our method. Given an input image (left), METRO takes non-local interactions among joints and vertices into consideration for human mesh reconstruction (right). We visualize the self-attentions between a specified joint and all other vertices, where brighter color indicates stronger attention. We observe that METRO discovers rich, input-dependent interactions among the joints and vertices.}
} 
\vspace{-2mm}
\label{fig:vis_3dpw_new_att}
\end{center}
\end{figure*}

\subsection{Main Results}

We compare METRO with the previous state-of-the-art methods on 3DPW and Human3.6M datasets. Following the literature~\cite{kocabas2019vibe,kolotouros2019learning,kanazawa2018end,kolotouros2019convolutional}, we conduct mix-training using 3D and 2D training data. 
The results are shown in Table~\ref{tbl:compare-h36m-3dpw}. 
Our method outperforms prior works on both datasets. 

First of all, we are interested in how transformer works for in-the-wild reconstruction of 3DPW. 
As shown in the left three columns of Table~\ref{tbl:compare-h36m-3dpw}, our method outperforms VIBE~\cite{kocabas2019vibe}, which was the state-of-the-art method on this dataset. It is worth noting that, VIBE is a video-based approach, whereas our method is an image-based approach. 

In addition, we evaluate the performance on the in-door scenario of Human3.6M. We follow the setting in the prior arts~\cite{kolotouros2019learning,Moon_2020_ECCV_I2L-MeshNet}, and train our model without using 3DPW data. The results are shown in the right two columns of Table~\ref{tbl:compare-h36m-3dpw}. Our method achieves better reconstruction performance, especially on PA-MPJPE metric.

The two datasets Human3.6M and 3DPW have different challenges. The scenes in 3DPW have more severe occlusions. The scenes in Human3.6 are simpler and the challenge is more on how to accurately estimate body shape. The fact that METRO works well on both datasets demonstrates that it is both robust to occlusions and capable of accurate body shape regression.

\subsection{Ablation Study}

\Paragraph{\textbf{Effectiveness of Masked Vertex Modeling}:} Since we design a Masked Vertex Modeling objective for transformer, one interesting question is whether the objective is useful. Table~\ref{tbl:compare-mpm} shows the ablation study on Human3.6M. We observe that Masked Vertex Modeling significantly improves the results. Moreover, we study how many percentage of query tokens should be masked. We vary the maximum masking percentage, and Table~\ref{tbl:compare-mpm-numbers} shows the comparison.
As we increase the number of masked queries for training, it improves the performance. However, the impact becomes less prominent if we mask more than $30\%$ of input queries. This is because large numbers of missing queries would make the training more difficult.

\begin{table}
\centering
\begin{tabular}{lcc}
    \toprule
	 & MPJPE $\downarrow$ & PA-MPJPE $\downarrow$ \\
	\midrule
     w/o MVM  & $61.0$ & $39.1$\\ 
     w/ MVM  & $\kevin{\textbf{54.0}}$ & $\kevin{\textbf{36.7}}$\\ 
	\bottomrule
\end{tabular}
\caption{Ablation study of the Masked Vertex Modeling (MVM) objective, evaluated on Human3.6M.}
\label{tbl:compare-mpm}
\end{table}

\begin{table}
\centering
\resizebox{1.\columnwidth}{!}{
    \begin{tabular}{lcccccc}
	\toprule
	 Max Percentage & $0\%$ & $10\%$ & $20\%$ & $30\%$ & $40\%$ & $50\%$\\
	 \midrule
	 PA-MPJPE & $39.1$ & $37.6$ & $37.5$  & $\textbf{36.7}$ & $38.2$ & $37.3$\\  
	\bottomrule
	\end{tabular}
}	
\caption{Ablation study of the Masked Vertex Modeling objective using different percentages of masked queries, evaluated on Human3.6M. The variable $n\%$ indicates we mask randomly from $0\%$ to $n\%$ of input queries.}
\label{tbl:compare-mpm-numbers}
\end{table}

\begin{table*}
\centering
\begin{tabular}{lcccc}
    \toprule
	Method  & PA-MPVPE $\downarrow$ & PA-MPJPE $\downarrow$ & F@5 mm $\uparrow$ & F@15 mm $\uparrow$\\
	\midrule
	Hasson et al~\cite{kanazawa2018end} & $13.2$ & $-$ & $0.436$ & $0.908$\\
	Boukhayma et al.~\cite{kolotouros2019convolutional} & $13.0$ & $-$ & $0.435$ & $0.898$\\
	FreiHAND~\cite{kolotouros2019learning} & $10.7$ & $-$ & $0.529$ & $0.935$\\
	Pose2Mesh~\cite{Choi_2020_ECCV_Pose2Mesh} & $7.8$ & $7.7$ & $0.674$ & $0.969$\\
	I2LMeshNet~\cite{Moon_2020_ECCV_I2L-MeshNet} & $7.6$ & $7.4$ & $0.681$ & $0.973$\\
	\midrule
    METRO (Ours)  & $\textbf{6.3}$ & $\textbf{6.5}$ & $\textbf{0.731}$ & $\textbf{0.984}$\\
	\bottomrule
\end{tabular}
\caption{Performance comparison with the state-of-the-art methods, evaluated on FreiHAND online server. METRO outperforms previous state-of-the-art approaches by a large margin. 
}
\label{tbl:compare-hand}
\end{table*}

\Paragraph{\textbf{Non-local Interactions}:} 
To further understand the effect of METRO in learning interactions among joints and mesh vertices, we conduct analysis on the self-attentions in our transformer. 

\kevin{Figure~\ref{fig:vis_3dpw_new_att} shows the visualization of the self-attentions and mesh reconstruction. For each row in Figure~\ref{fig:vis_3dpw_new_att}, we show the input image, and the self-attentions between a specified joint and all the mesh vertices. The brighter color indicates stronger attention. At the first row, the subject is severely occluded and the right body parts are invisible. As we predict the location of right wrist, METRO attends to relevant non-local vertices, especially those on the head and left hand. 
At the bottom row, the subject is heavily bended. For the head position prediction, METRO attends to the feet and hands (6th column at the bottom row).  It makes sense intuitively since the hand and foot positions provide strong cues to the body pose and subsequently the head position.
Moreover, we observe the model performs self-attentions in condition to the input image. As shown in the second row of Figure~\ref{fig:vis_3dpw_new_att}, when predicting the location of right wrist, METRO focuses more on the right foot which is different from the attentions in the other three rows. }

\begin{figure}
\begin{center}
\includegraphics[trim=0 0 00 00, clip,width=1\columnwidth]{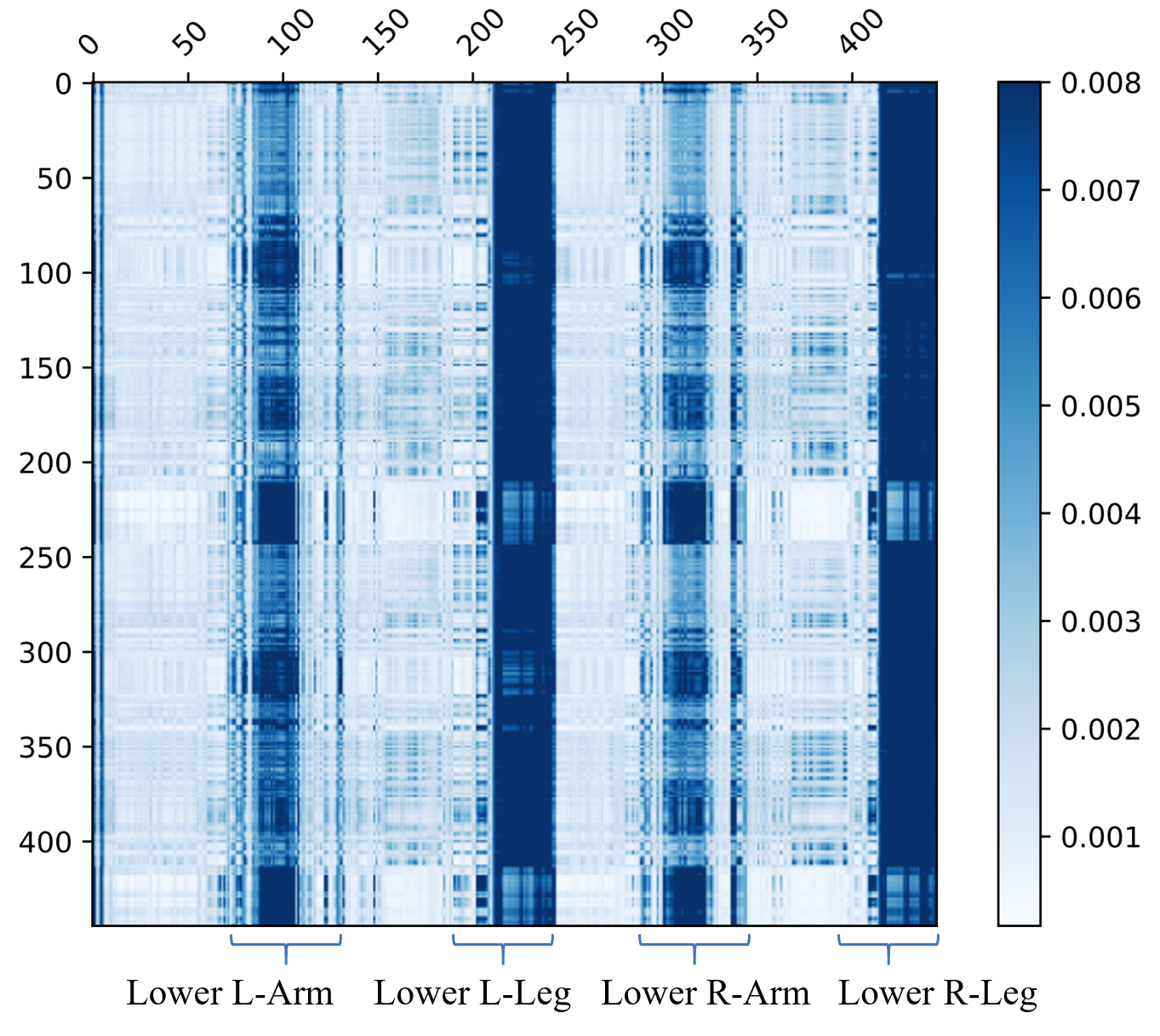}
\caption{
Visualization of self-attentions among body joints and mesh vertices. The x-axis and y-axis correspond to the queries and the predicted outputs, respectively. The first 14 columns from the left correspond to the body joints. The rest of columns correspond to the mesh vertices. \kevin{Each row shows the attention weight $w_{i,j}$ of the $j$-th query for the $i$-th output. Darker color indicates stronger attention.}
} 
\vspace{-0mm}
\label{fig:att_joint_vertices}
\end{center}
\end{figure}

\kevin{We further conduct quantitative analysis on the non-local interactions.}
We randomly sample $5000$ images from 3DPW test set, and estimate an overall self-attention map. \camready{It is the average attention weight of all attention heads at the last transformer layer.} \kevin{We visualize the interactions among $14$ body joints and $431$ mesh vertices in Figure~\ref{fig:att_joint_vertices}. Each pixel shows the intensity of self-attention, where darker color indicates stronger attention.
Note that the first $14$ columns are the body joints, and the rest of them represent the mesh vertices. We observe that METRO pays strong attentions to the vertices on the lower arms and the lower legs. 
This is consistent with the inverse kinematics literature~\cite{aristidou2018inverse} where the interior joints of a linked figure can be estimated from the position of an end effector. 
}

\begin{table}
\centering
\begin{tabular}{lcc}
    \toprule
	Backbone &  MPJPE $\downarrow$ & PA-MPJPE $\downarrow$ \\
	\midrule
     \kevin{ResNet50} & $56.5$ & $40.6$\\
     \kevin{HRNet-W40} & $55.9$ & $38.5$\\
    \kevin{HRNet-W64} & $\kevin{\textbf{54.0}}$ & $\kevin{\textbf{36.7}}$\\
	\bottomrule
\end{tabular}
\caption{\camready{Analysis on different backbones,} evaluated on Human3.6M. All backbones are pre-trained on ImageNet. We observe that increasing the number of filters in the high resolution feature maps of HRNet is beneficial to mesh regression.}
\label{tbl:compare-input-feat}
\end{table}

\begin{figure*}[h]
\begin{center}
\includegraphics[trim=0 0 0 0, clip,width=1\textwidth]{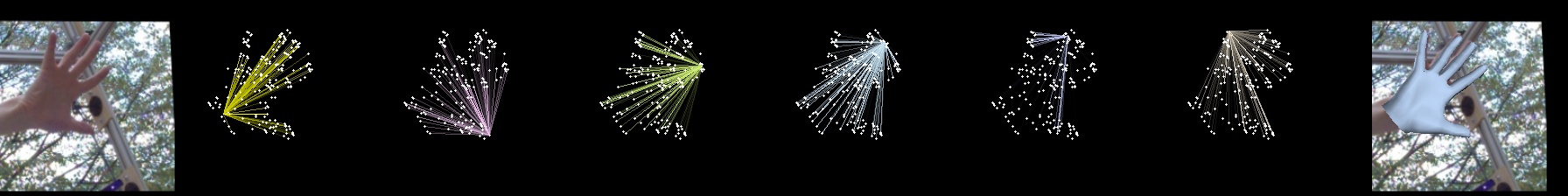}\\
\includegraphics[trim=0 0 0 0, clip,width=1\textwidth]{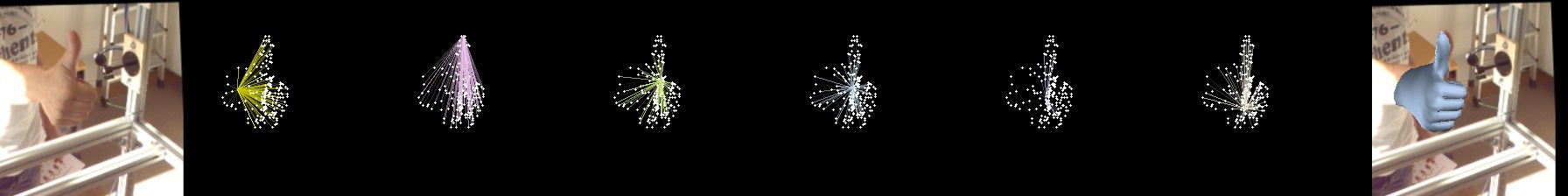}\\
\includegraphics[trim=0 0 0 0, clip,width=1\textwidth]{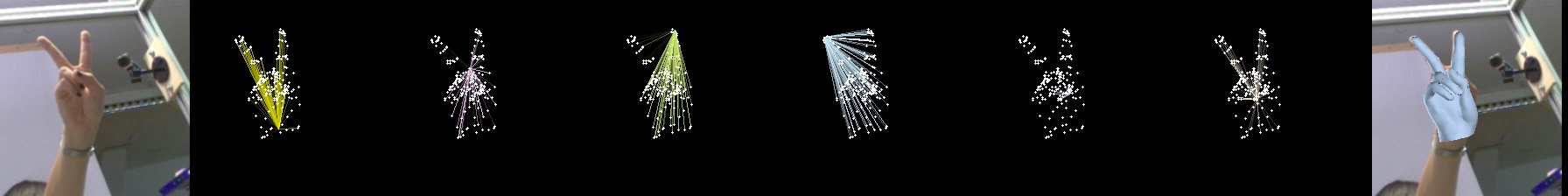}\\
\includegraphics[trim=0 0 0 0, clip,width=1\textwidth]{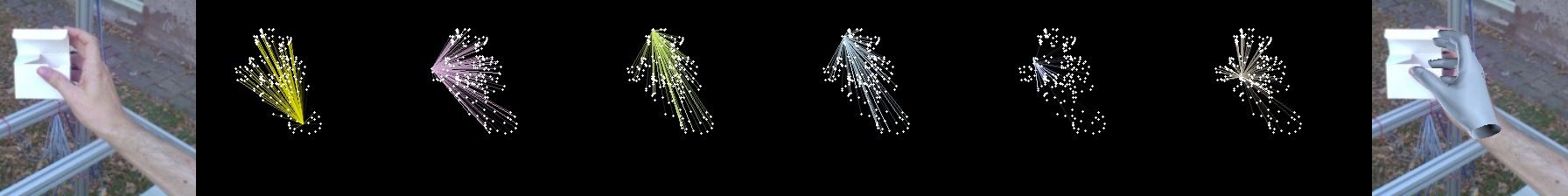}\\
\includegraphics[trim=0 0 0 0, clip,width=1\textwidth]{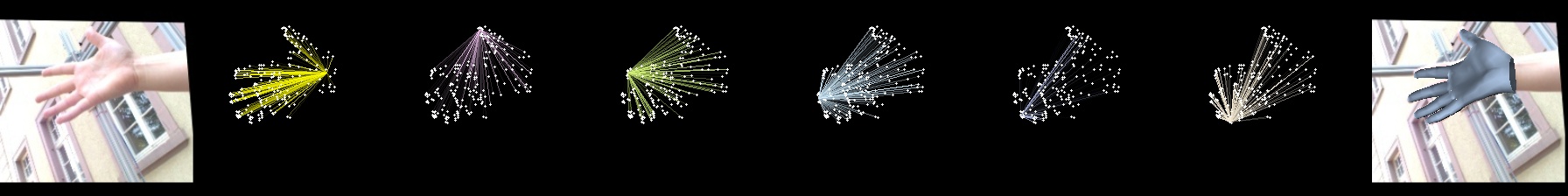}\\
\setlength{\tabcolsep}{19pt}
\begin{tabular}{cccccccc}
Input & Wrist &  Thumb &  Index &  Middle & Ring & Pinky & Output
\end{tabular}
\vspace{-2mm}
\caption{
\kevin{Qualitative results of our method on FreiHAND test set. We visualize the self-attentions between a specified joint and all the mesh vertices, where brighter color indicates stronger attention. METRO is a versatile framework that can be easily extended to 3D hand reconstruction.}
} 
\vspace{-4mm}
\label{fig:vis_hand_att}
\end{center}
\end{figure*}

\Paragraph{\textbf{Input Representations}:} We study the behaviour of our transformer architecture by using different CNN backbones. We use ResNet50~\cite{he2016deep} and HRNet~\cite{WangSCJDZLMTWLX19} variations for this experiment. All backbones are pre-trained on the $1000$-class image classification task of ImageNet~\cite{russakovsky2015imagenet}. For each backbone, we extract a global image feature vector $X\in \mathbb{R}^{2048\times1}$, and feed it into the transformer. In Table~\ref{tbl:compare-input-feat}, we observe our transformer achieves competitive performance when using a ResNet50 backbone. \camready{As we increase the channels of the high-resolution feature maps in HRNet,} we observe further improvement. 

\Paragraph{\textbf{Generalization to 3D Hand in-the-wild}:} METRO is capable of predicting arbitrary joints and vertices, without the dependencies on adjacency matrix and parametric coefficients. Thus, METRO is highly flexible and general for mesh reconstruction of other objects. 
To demonstrate this capability, we conduct experiment on FreiHAND~\cite{zimmermann2019freihand}.
We train our model on FreiHAND from scratch, and evaluate results on FreiHAND online server. Table~\ref{tbl:compare-hand} shows the comparison with the prior works. 
METRO outperforms previous state-of-the-art methods by a large margin. Without using any external training data, METRO achieved the first place on FreiHAND leaderboard \camready{at the time of paper submission}\footnote{According to the official FreiHAND leaderboard in November 2020: \href{https://competitions.codalab.org/competitions/21238}{https://competitions.codalab.org/competitions/21238}}.

Figure~\ref{fig:vis_hand_att} shows our qualitative results with non-local interactions. \kevinarxiv{In the appendix, we provide further analysis on the 3D hand joints, and show that the self-attentions learned in METRO are consistent with inverse kinematics~\cite{aristidou2018inverse}. }

\pdfoutput=1
\section{Conclusion}\label{sec:conclude}
We present a simple yet effective mesh transformer framework to reconstruct human pose and mesh from a single input image. 
We propose the Masked Vertex Modeling objective to learn non-local interactions among body joints and mesh vertices. Experimental results show that, our method advances the state-of-the-art performance on 3DPW, Human3.6M, \kevinarxiv{and FreiHAND datasets.}

A detailed analysis reveals that the performance improvements are mainly attributed to the input-dependent non-local interactions learned in METRO, which enables predictions based on important joints and vertices, regardless of the mesh topology. We further demonstrate the generalization capability of the proposed approach to 3D hand reconstruction.

{\small
\bibliographystyle{ieee_fullname}
\bibliography{humanmesh}
}

\clearpage
\appendix
\pdfoutput=1

\twocolumn[{%
\renewcommand\twocolumn[1][]{#1}%
\begin{center}
\textbf{\large APPENDICES}
\end{center}
\vspace{2mm}
\begin{center}
\begin{tabular}{lcccccc}
    \toprule
    \multirow{1}{*}{} & \multicolumn{3}{c}{3DPW} & & \multicolumn{2}{c}{Human3.6M}\\ 
    \cline{2-4}\cline{6-7}
	Method  & MPVE $\downarrow$ & MPJPE $\downarrow$ & PA-MPJPE $\downarrow$ & &  MPJPE $\downarrow$ & PA-MPJPE $\downarrow$ \\
	\midrule
	HMR~\cite{kanazawa2018end} & $-$ & $-$ & $81.3$ && $88.0$ & $56.8$  \\
	GraphCMR~\cite{kolotouros2019convolutional} & $-$ & $-$ & $70.2$ && $-$ & $50.1$\\
	SPIN~\cite{kolotouros2019learning} & $116.4$ & $-$ & $59.2$ && $-$ & $41.1$\\
	Pose2Mesh~\cite{Choi_2020_ECCV_Pose2Mesh} & $-$ & $89.2$ & $58.9$ && $64.9$ & $47.0$\\
	I2LMeshNet~\cite{Moon_2020_ECCV_I2L-MeshNet} & $-$ & $93.2$ & $57.7$ && $55.7$ & $41.1$\\
	VIBE~\cite{kocabas2019vibe} & $99.1$ & $82.0$ & $51.9$ && $65.6$ & $41.4$\\
	\midrule
	HoloPose~\cite{guler2019holopose}  & $-$ & $-$ & $-$ && $60.2$ & $46.5$\\
	Arnab \textit{et al.}~\cite{arnab2019exploiting} & $-$ & $-$ & $72.2$ && $77.8$ & $54.3$\\
	DenseRaC~\cite{xu2019denserac} & $-$ & $-$ & $-$ && $-$ & $48.0$\\
	Zhang \textit{et al.}~\cite{zhang2020object} & $-$ & $-$ & $72.2$ && $-$ & $41.7$\\
	Zeng \textit{et al.}~\cite{zeng20203d} & $-$ & $-$ & $-$ && $60.6$ & $39.3$\\
	HKMR~\cite{georgakis2020hierarchical} & $-$ & $-$ & $-$ && $59.6$ & $43.2$\\
	\midrule
    METRO (Ours)  & $\textbf{88.2}$ & $\textbf{77.1}$ & $\textbf{47.9}$ && $\kevin{\textbf{54.0}}$ & $\kevin{\textbf{36.7}}$\\ 
	\bottomrule
\end{tabular}
\figcaption{Adding references (HKMR~\cite{georgakis2020hierarchical}, Zeng \textit{et al.}~\cite{zeng20203d}, Zhang \textit{et al.}~\cite{zhang2020object}, DenseRaC~\cite{xu2019denserac}, Arnab \textit{et al.}~\cite{arnab2019exploiting}, HoloPose~\cite{guler2019holopose}) to the comparisons on 3DPW and Human 3.6M datasets. \label{tbl:compare-h36m-3dpw_add}
}
\vspace{5mm}
\end{center}
}]

\label{sec:appendix}

\section{Additional Reference}

We would like to add additional references (HKMR~\cite{georgakis2020hierarchical}, Arnab \textit{et al.}~\cite{arnab2019exploiting}, Zeng \textit{et al.}~\cite{zeng20203d}, Zhang \textit{et al.}~\cite{zhang2020object}, DenseRaC~\cite{xu2019denserac}, HoloPose~\cite{guler2019holopose}). Among the new references, HKMR~\cite{georgakis2020hierarchical} regresses SMPL parameters by leveraging a pre-specified hierarchical kinematic structure that consists of a root chain and five child chains corresponding to 5 end effectors (head, left/right arms, left/right legs). HoloPose~\cite{guler2019holopose} estimates rotation angles of body joints, and uses it as the prior to guide part-based human mesh reconstruction. Zeng \textit{et al.}~\cite{zeng20203d} designs the continuous UV map to preserve neighboring relationships of the mesh vertices.  Zhang \textit{et al.}~\cite{zhang2020object} addresses the occlusion scenario by formulating the task as a UV-map inpainting problem. Since 3DPW is a relatively new benchmark, most literature reported results on Human3.6M, but not 3DPW. We have added their Human3.6M results in Table~\ref{tbl:compare-h36m-3dpw_add}. As we can see, our method outperforms all of the prior works by a large margin.

\camready{Recently, researchers are exploring the transformer models for other 3D vision topics, such as multi-view human pose estimation~\cite{he2020epipolar} and hand pose estimation based on point could~\cite{huang2020hand}. We encourage the readers to undertake these studies for further explorations.}

\begin{table*}
\centering
\begin{tabular}{lccc}
    \toprule
	Model & Dimensionality Reduction Scheme & PA-MPJPE $\downarrow$ \\
	\midrule
    Transformer~\cite{vaswani2017attention} & $(H+3)$ $\rightarrow$ 3 & $208.7$\\  
    METRO & $(H+3)$ $\rightarrow$ $H/2$ $\rightarrow$ 3 &  $192.1$\\
    METRO & $(H+3)$ $\rightarrow$ $H/2$ $\rightarrow$ $H/4$ $\rightarrow$ 3 &  $43.8$\\  
    METRO & $(H+3)$ $\rightarrow$ $H/2$ $\rightarrow$ $H/4$ $\rightarrow$ $H/8$ $\rightarrow$ 3 & $\textbf{36.7}$\\  
	\bottomrule
\end{tabular}
\caption{Performance comparison of different dimentionality reduction schemes, evaluated on Human3.6M validation set. Please note that all the transformer variants have the same total number of hidden layers (12 layers) for fair comparison. $H$=2048.} 
\label{tbl:abl-progress}
\vspace{0mm}
\end{table*}

\begin{table}
\centering
\begin{tabular}{lcc}
    \toprule
	 & CNN (HRNet-W64) & Transformer\\
	\midrule
	\# Parameters & 128M & \textbf{102M} \\ 
	Inference time & 52.05 ms & \textbf{28.22 ms}\\
	\bottomrule
\end{tabular}
\caption{Number of parameters and inference time per image. The runtime speed is estimated by using batch size 1.}
\label{tbl:complexity}
\end{table}

\camready{
\section{Implementation Details and Computation Resource}
We develop METRO using PyTorch and Huggingface transformer library. We conduct training on a machine equipped with $8$ NVIDIA V100 GPUs (32GB RAM) and we use batch size $32$. Each epoch takes 32 minutes and we train for 200 epochs. Overall, our training takes 5 days. We use the Adam optimizer and a step-wise learning rate decay. We set the initial learning rate as $1\times10^{-4}$ for both transformer and CNN backbone. The learning rate is decayed by a factor of $10$ at the $100$th epoch. Our multi-layer transformer encoder is randomly initialized, and the CNN backbone is initialized with ImageNet pre-trained weights. Following~\cite{kolotouros2019convolutional,kolotouros2019learning}, we apply standard data augmentation during training.}

\camready{
We evaluate the runtime speed of our model using a machine equipped with a single NVIDIA P100 GPU (16GB RAM). Our runtime speed is about 12 fps using batch size 1. The runtime speed can be accelerated to around 24 fps using batch size 32. Table~\ref{tbl:complexity} shows the details of each module in METRO.
}

\camready{
For our masked vertex modeling, following BERT~\cite{devlin2019bert}, we implement it by using a pre-defined special [MASK] token (2051-D floating value vector in our case) to replace the randomly selected input queries.}

\begin{table}
\centering
\begin{tabular}{lc}
    \toprule
	Positional Encoding & PA-MPJPE $\downarrow$ \\
	\midrule
	Sinusoidal~\cite{vaswani2017attention}   & $37.5$\\ 
    Ours & $\textbf{36.7}$\\  
	\bottomrule
\end{tabular}
\caption{Comparison of different positional encoding schemes, evaluated on Human3.6M validation set.} 
\label{tbl:abl-pos-enc}
\vspace{0mm}
\end{table}

\camready{
\Section{Progressive Dimensionality Reduction} 
Since we gradually reduce the hidden sizes in the transformer architecture, one interesting question is whether such a progressive dimensionality reduction scheme is useful. We have conducted an ablation study on different schemes, and Table~\ref{tbl:abl-progress} shows the comparison. 
In Table~\ref{tbl:abl-progress}, the row ``(H+3)$\rightarrow$3" corresponds to a baseline using one linear projection $H+3$ to $3$. The result is poor. Row ``(H+3)$\rightarrow$H/2$\rightarrow$3" is another baseline which keeps a smaller dimension throughout the network. The result is also bad. Our finding is that large-step (steep) dimension reduction does not work well for 3D mesh regression. Our progressive scheme is inspired by~\cite{hinton2006reducing} which performed dimensionality reduction gradually with multiple blocks.
}

\camready{
\Section{Positional Encoding} 
Since our positional encoding is different from the conventional one, one may wonder what if we use sinusoidal functions~\cite{vaswani2017attention} but not a template mesh. We have compared our method with the conventional positional encoding which uses sinusoidal functions, and Table~\ref{tbl:abl-pos-enc} shows the results. We see that using sinusoidal functions is slightly worse. This is probably because directly encoding coordinates makes it more efficient to learn 3D coordinate regression.
}

\section{Qualitative Results}

Figure~\ref{fig:vis_3dpw_challeging} shows a qualitative comparison with the previous image-based state-of-the-art methods~\cite{Moon_2020_ECCV_I2L-MeshNet,kolotouros2019convolutional} in challenging scenarios. These methods only use a single frame as input. In the first row, the subject is heavily bending. Prior works have difficulty in reconstructing a correct body shape for the subject. In contrast, our method reconstructs a reasonable human mesh with correct pose. In the second row, the subject is occluded by the vehicle. We see that prior works are sensitive to the occlusions, and failed to generate correct human mesh. In contrast, our method performs more robustly in this occlusion scenario. In the bottom row, the subject is sitting on the chair. Our method reconstructed a better human mesh compared to the previous state-of-the-art methods. 

Figure~\ref{fig:vis_freihand} shows the qualitative results of our method on 3D hand reconstruction. Without making any modifications to the network architecture, our method works well for hands and is robust to occlusions. It demonstrates our method's advantage that it can be easily extended to other types of objects.

\begin{figure}
\begin{center}
(a) Example1
\includegraphics[width=1.\columnwidth]{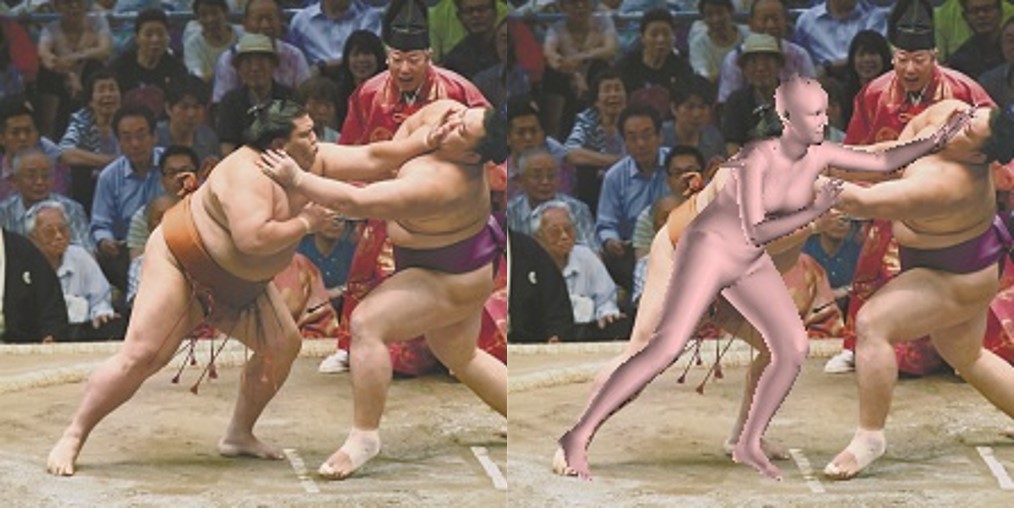}\\
\setlength{\tabcolsep}{45.0pt}
\begin{tabular}{cc}
Input & Output\\ \\
\end{tabular}
(b) Example2\\
\includegraphics[width=1.\columnwidth]{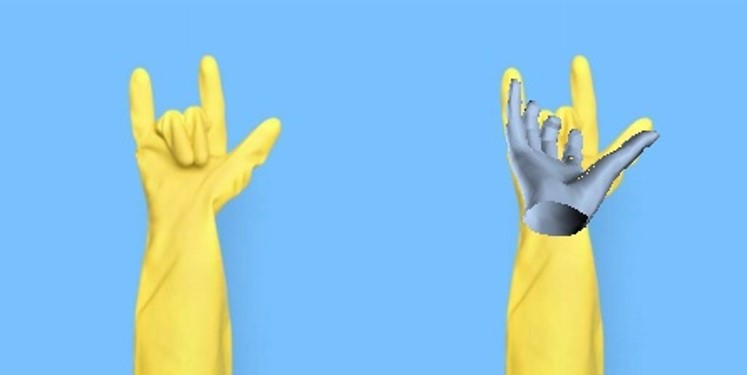}\\
\setlength{\tabcolsep}{45.0pt}
\begin{tabular}{cc}
Input & Output\\ \\
\end{tabular}
(c) Example3\\
\includegraphics[width=1.\columnwidth]{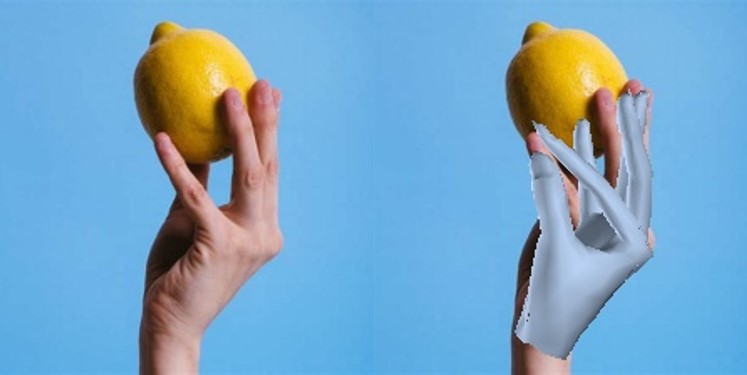}\\
\setlength{\tabcolsep}{45.0pt}
\begin{tabular}{cc}
Input & Output\\
\end{tabular}
\caption{
Failure cases. METRO may not perform well when the testing sample is very different from the training data.
} 
\vspace{-0mm}
\label{fig:limitation}
\end{center}
\end{figure}

\begin{table*}
\centering
\begin{tabular}{lcccc}
    \toprule
	Method  & PA-MPVPE $\downarrow$ & PA-MPJPE $\downarrow$ & F@5 mm $\uparrow$ & F@15 mm $\uparrow$\\
	\midrule
	I2LMeshNet~\cite{Moon_2020_ECCV_I2L-MeshNet} & $7.6$ & $7.4$ & $0.681$ & $0.973$\\
	\midrule
    METRO  & $6.7$ & $6.8$ & $0.717$ & $0.981$\\
    METRO + Test time augmentation  & $\textbf{6.3}$ & $\textbf{6.5}$ & $\textbf{0.731}$ & $\textbf{0.984}$\\
	\bottomrule
\end{tabular}
\caption{Effectiveness of test-time augmentation on FreiHAND test set.
}
\label{tbl:compare-hand-tta}
\end{table*}

\section{Non-local Interactions of Hand Joints}
We further conduct quantitative analysis on the non-local interactions among hand joints learned by our model.
We randomly sample $1000$ samples from FreiHAND test set, and estimate an overall self-attention map. Figure~\ref{fig:att_joint_vertices_hand} shows the interactions among $21$ hand joints. 
There are 21 rows and 21 columns. Pixel ($i$, $j$) represents the amount of attention that hand joint $i$ attends to joint $j$. A darker color indicates stronger attention. We can see that the wrist joint (column 0) receives strong attentions from all the joints. Intuitively wrist joint acts like a ``root" of the hand’s kinematics tree. In addition, columns 4, 8, 12, and 16 receive strong attentions from many other joints. These columns correspond to the tips of thumb, index, middle, and ring fingers, respectively. These finger tips are end effectors~\cite{aristidou2018inverse} and they can be used to estimate the interior joint positions in inverse kinematics. On the other hand, the tip of pinky only receives attentions from the joints on the ring finger. This is probably because pinky is not as active as the other fingers and its motion is more correlated to the ring finger compared to the other fingers.

\camready{
\section{Test-Time Augmentation for FreiHAND}
We have explored test-time augmentation in our FreiHAND experiments. We do not use test-time augmentation in Human3.6M and 3DPW experiments. Given a test image, we apply different rotations and scaling to the test image. We then feed these transformed images to our model, and average the results to obtain the final output mesh. In order to compute an average 3D mesh, we perform 3D alignment (i.e., Procrustes analysis~\cite{gower1975generalized}) to normalize the output meshes. In Table~\ref{tbl:compare-hand-tta}, we empirically observed that such an implementation is helpful to improve $0.4$ PA-MPVPE on FreiHAND test set.}

\camready{
\Section{Limitations}
As METRO is a data-driven approach, it may not perform well when the testing sample is very different from the training data. We show some example failure cases in Figure~\ref{fig:limitation} where the test images are downloaded from the Internet. First, as shown in Figure~\ref{fig:limitation}(a), we observed that if the target body shape is very different from the existing training data (i.e., SMPL style data), our method may not faithfully reconstruct the muscles of the subject. Secondly, as shown in Figure~\ref{fig:limitation}(b), our model fails to reconstruct a correct mesh due to the fact that there is no glove data in the training set. Finally, the proposed method is a mesh-specific approach. If we were to apply our pre-trained right-hand model to the left-hand images, as can be seen in Figure~\ref{fig:limitation}(c), our model will not work well. How to develop a unified model for different 3D objects is an interesting future work.}

\begin{figure*}[h]
\begin{center}
\includegraphics[trim=0 0 0 0, clip,width=0.8\textwidth]{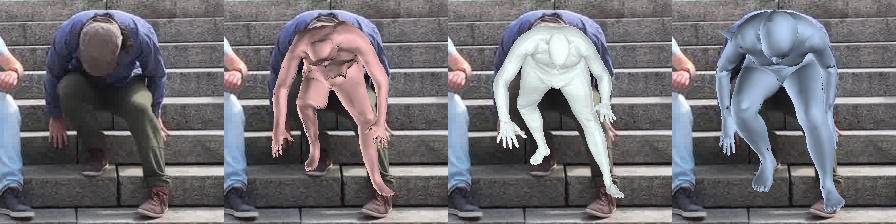}\\
\includegraphics[trim=0 0 0 0, clip,width=0.8\textwidth]{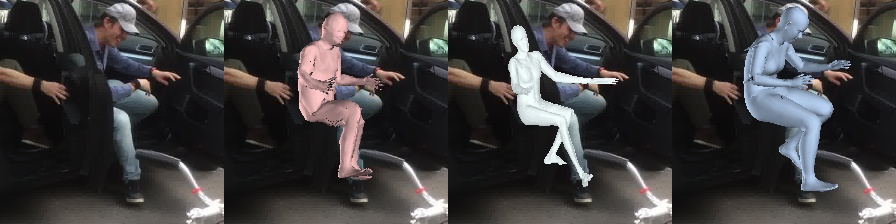}\\
\includegraphics[trim=0 0 0 0, clip,width=0.8\textwidth]{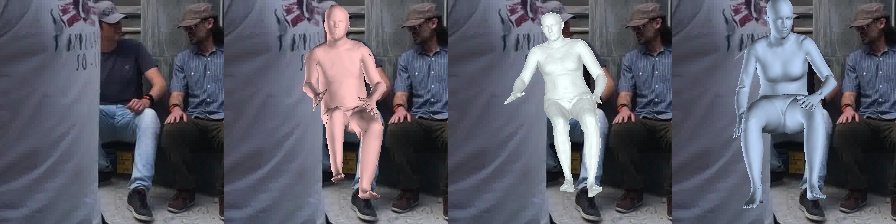}\\
\includegraphics[trim=0 0 0 0, clip,width=0.8\textwidth]{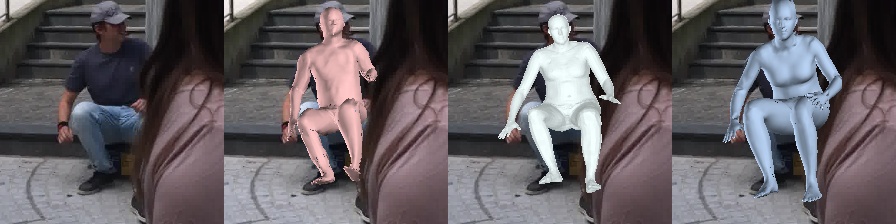}\\
\includegraphics[trim=0 0 0 0, clip,width=0.8\textwidth]{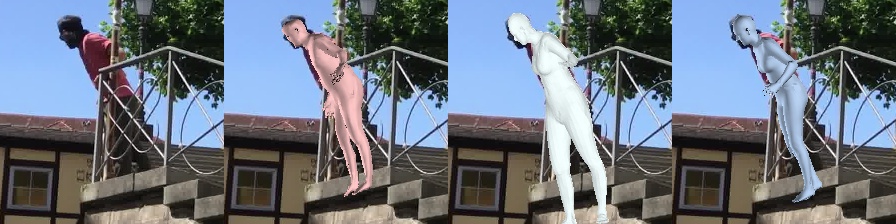}\\
\includegraphics[trim=0 0 0 0, clip,width=0.8\textwidth]{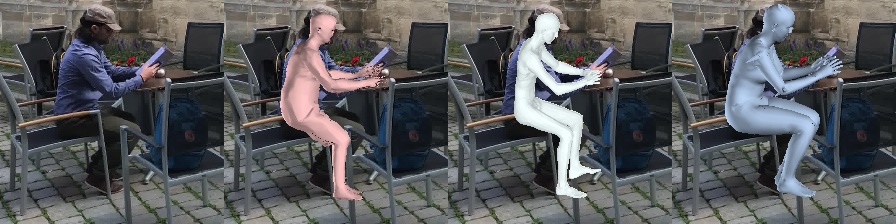}\\
\setlength{\tabcolsep}{30pt}
\begin{tabular}{cccc}
\ \ \ \ \ Input &  GraphCMR~\cite{kolotouros2019convolutional} &  I2L-M~\cite{Moon_2020_ECCV_I2L-MeshNet} &  Ours
\end{tabular}
\vspace{-2mm}
\caption{
Qualitative comparison between our method and other single-frame-based approaches. Our method is more robust to challenging poses and occlusions.   
} 
\label{fig:vis_3dpw_challeging}
\end{center}
\end{figure*}

\begin{figure*}
\begin{center}
\includegraphics[trim=0 0 0 0, clip,width=0.33\textwidth]{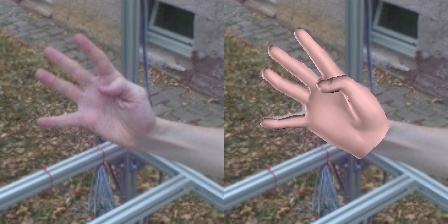}
\includegraphics[trim=0 0 0 0, clip,width=0.33\textwidth]{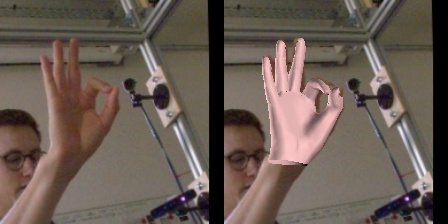}
\includegraphics[trim=0 0 0 0, clip,width=0.33\textwidth]{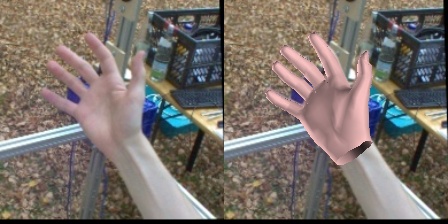}\\
\includegraphics[trim=0 0 0 0, clip,width=0.33\textwidth]{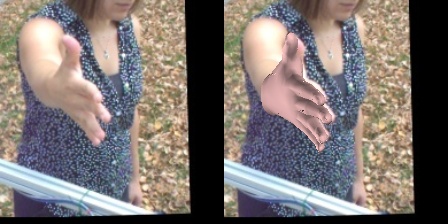}
\includegraphics[trim=0 0 0 0, clip,width=0.33\textwidth]{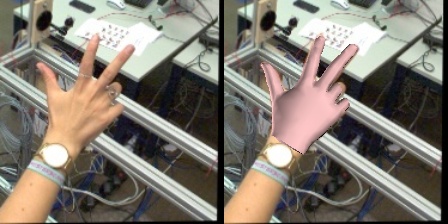}
\includegraphics[trim=0 0 0 0, clip,width=0.33\textwidth]{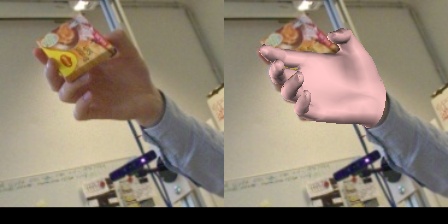}\\
\includegraphics[trim=0 0 0 0, clip,width=0.33\textwidth]{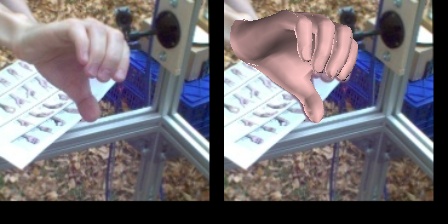}
\includegraphics[trim=0 0 0 0, clip,width=0.33\textwidth]{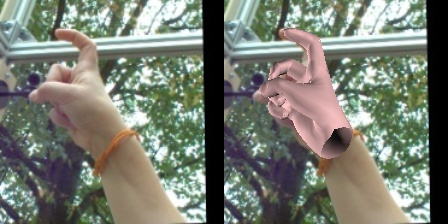}
\includegraphics[trim=0 0 0 0, clip,width=0.33\textwidth]{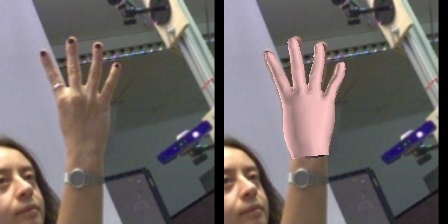}\\
\includegraphics[trim=0 0 0 0, clip,width=0.33\textwidth]{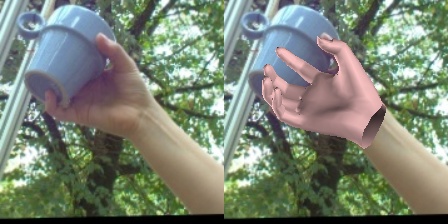}
\includegraphics[trim=0 0 0 0, clip,width=0.33\textwidth]{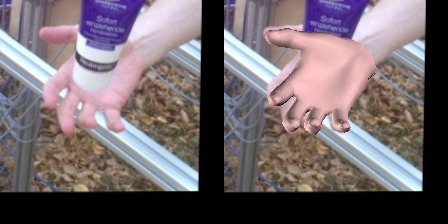}
\includegraphics[trim=0 0 0 0, clip,width=0.33\textwidth]{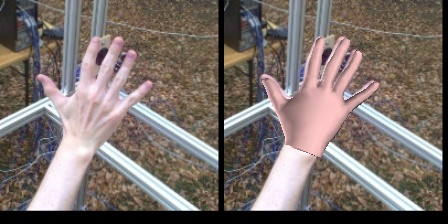}\\
\includegraphics[trim=0 0 0 0, clip,width=0.33\textwidth]{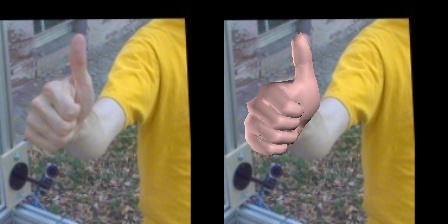}
\includegraphics[trim=0 0 0 0, clip,width=0.33\textwidth]{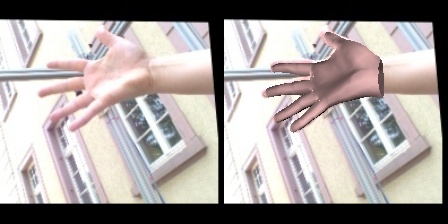}
\includegraphics[trim=0 0 0 0, clip,width=0.33\textwidth]{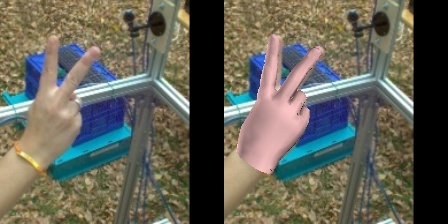}\\
\caption{
Qualitative results of our method on FreiHAND test set. Our method can be easily extended to reconstruct 3D hand mesh.
} 
\label{fig:vis_freihand}
\end{center}
\end{figure*}

\begin{figure*}
\begin{center}
\includegraphics[trim=0 0 00 00, clip,width=1\textwidth]{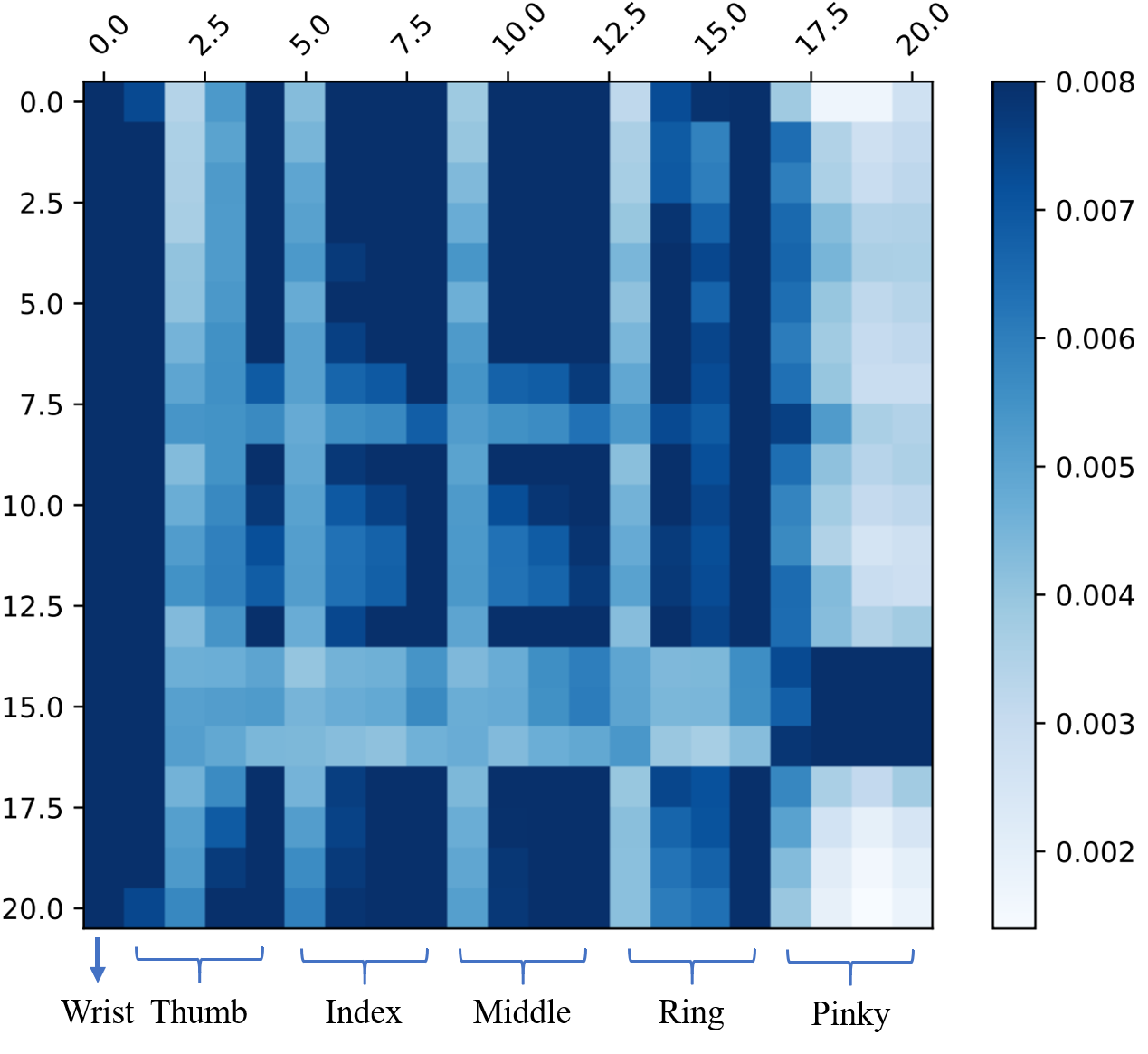}
\caption{
Visualization of self-attentions among hand joints. There are 21 rows and 21 columns corresponding to 21 hand joints. Pixel ($i$, $j$) represents the amount of attention that joint $i$ attends to joint $j$. A darker color indicates stronger attention. The definition of the 21 joints is shown in Figure~\ref{fig:hand_joint_define}. 
} 
\vspace{-0mm}
\label{fig:att_joint_vertices_hand}
\end{center}
\end{figure*}

\begin{figure*}
\begin{center}
\includegraphics[trim=0 0 00 00, clip,width=0.7\textwidth]{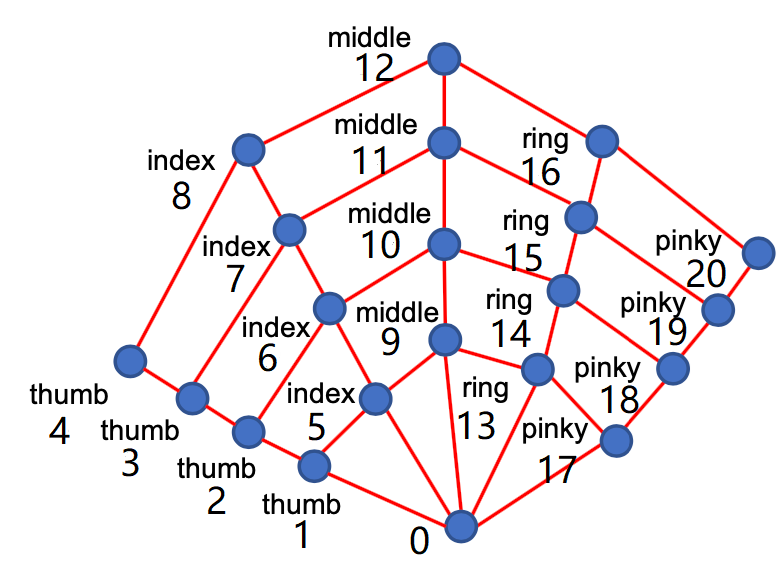}
\caption{
Definition of the hand joints. The illustration is adapted from~\cite{Choi_2020_ECCV_Pose2Mesh}.
} 
\vspace{-0mm}
\label{fig:hand_joint_define}
\end{center}
\end{figure*}

\end{document}